\pdfoutput=1
\documentclass[11pt]{article}

\usepackage[]{emnlp2021}

\usepackage{times}
\usepackage{latexsym}
\usepackage[T1]{fontenc}
\usepackage[utf8]{inputenc}
\usepackage{microtype}
\usepackage{xcolor}
\usepackage{xspace}

\newcommand{\myparagraph}[1]{\paragraph{#1}}

\usepackage{subfig}
\usepackage{graphicx}
\usepackage{booktabs}
\usepackage{tabularx}
\usepackage{enumitem}
\usepackage{tcolorbox}
\usepackage{multicol}

\graphicspath{{./figures/}}

\newcommand{\missingcitation}[1]{\textcolor{magenta}{[CITE]}}

\newcommand{\Sref}[1]{\S\ref{#1}}

\newcommand{\Fref}[1]{Figure~\ref{#1}}

\newcommand{\Tref}[1]{Table~\ref{#1}}
\newcommand{\Aref}[1]{Appendix~\ref{#1}}

\usepackage{xcolor}

\newcommand{\data}{\textsc{NormVio}\xspace}
\newcommand{\majority}{\textsc{\colorbox{blue!25}{Majority}}\xspace}
\newcommand{\perspective}{\textsc{\colorbox{orange!30}{Perspective}}\xspace}
\newcommand{\incivil}{\textsc{\colorbox{green!20}{IncivilHate}}\xspace}
\newcommand{\finalcomment}{\textsc{\colorbox{red!40}{Comment}}\xspace}
\newcommand{\history}{\textsc{\colorbox{violet!30}{+History}}\xspace}
\newcommand{\community}{\textsc{\colorbox{brown!20}{+Community}}\xspace}
\newcommand{\histcommunity}{\textsc{\colorbox{purple!25}{+History+Community}}\xspace}
\newcommand{\ruletext}{\textsc{\colorbox{darkgray!20}{+Rule}}\xspace}
\newcommand{\rulehist}{\textsc{\colorbox{olive!30}{+Rule+History}}\xspace}
\newcommand{\rulecommunity}{\textsc{\colorbox{cyan!20}{+Rule+History+Community}}\xspace}

\title{Detecting Community Sensitive Norm Violations in Online Conversations}

\author{
  Chan Young Park$^\clubsuit$ \quad Julia Mendelsohn$^\spadesuit$ \quad Karthik Radhakrishnan$^\clubsuit$ \quad Kinjal Jain$^\clubsuit$\\
  \textbf{Tushar Kanakagiri}$^\clubsuit$ \quad \textbf{David Jurgens}$^\spadesuit$ \quad \textbf{Yulia Tsvetkov}$^\heartsuit$\\
  \vspace{0.5mm}$^\clubsuit$Language Technologies Institute, Carnegie Mellon University \\
$^\spadesuit$ University of Michigan \\
$^\heartsuit$Paul G.~Allen School of Computer Science \& Engineering, University of Washington \\
        {\tt \{chanyoun,kradhak2,kinjalj,tkanakag\}@cs.cmu.edu,}\\
        {\tt\{juliame,jurgens@umich.edu\}, yuliats@cs.washington.edu
        }
}

\begin{document}
\maketitle
\begin{abstract}

Online platforms and communities establish their own norms that govern what behavior is acceptable within the community. Substantial effort in NLP has focused on identifying unacceptable behaviors and, recently, on forecasting them before they occur. However, these efforts have largely focused on toxicity as the sole form of community norm violation. Such focus has overlooked the much larger set of rules that moderators enforce. Here, we introduce a new dataset focusing on a more complete spectrum of community norms and their violations in the local conversational and global community contexts.
We introduce a series of models that use this data to develop context- and community-sensitive norm violation detection, showing that these changes give high performance.\footnote{Dataset, code, and models are publicly available at \href{https://github.com/chan0park/NormVio}{https://github.com/chan0park/NormVio}.}
\end{abstract}

\section{Introduction}

Online communities establish their own norms of what is acceptable behavior \citep{danescu2013no,jhaver2018online,rajadesingan2020quick}. These norms run the gamut from \textit{no hate speech} or \textit{no personal attacks} to more idiosyncratic expectations of \textit{content formatting} and \textit{content sharing} \citep{chandrasekharan2018internet, Fiesler_Jiang_McCann_Frye_Brubaker_2018}. 
Community moderators are responsible for identifying and removing rule-breaking content, regardless of whether users violate rules intentionally or unintentionally due to unfamiliarity with community norms. 

Moderators of online communities often face a tough challenge of triaging the massive flow of content \citep{kiene2016surviving,dosono2019moderation, kiene2019technological}; for example, over 2 billion comments were posted to Reddit in just 2020.\footnote{\href{https://backlinko.com/reddit-users\#reddit-statistics}{https://backlinko.com/reddit-users\#reddit-statistics}}
Moderators have looked to technology to help support their role, using regex-based tools like Automoderator to flag potentially rule-breaking comments \citep{jhaver2019human}. Prior work has aimed to assist by developing machine learning techniques to recognize unacceptable content---yet these have focused on only the most socially-harmful violations, such as hate speech.
Furthermore, the rules moderators enforce vary widely both in their formulation and interpretation across communities, making a one-size-fits-all approach increasingly brittle.  
Since successful moderation relies on fine-grained understanding of a given community's norms, we present a new dataset and models for community-specific, contextualized norm violation detection for over twenty types of norms.

\begin{figure}[t!]
  \centering
  \includegraphics[width=0.7\columnwidth]{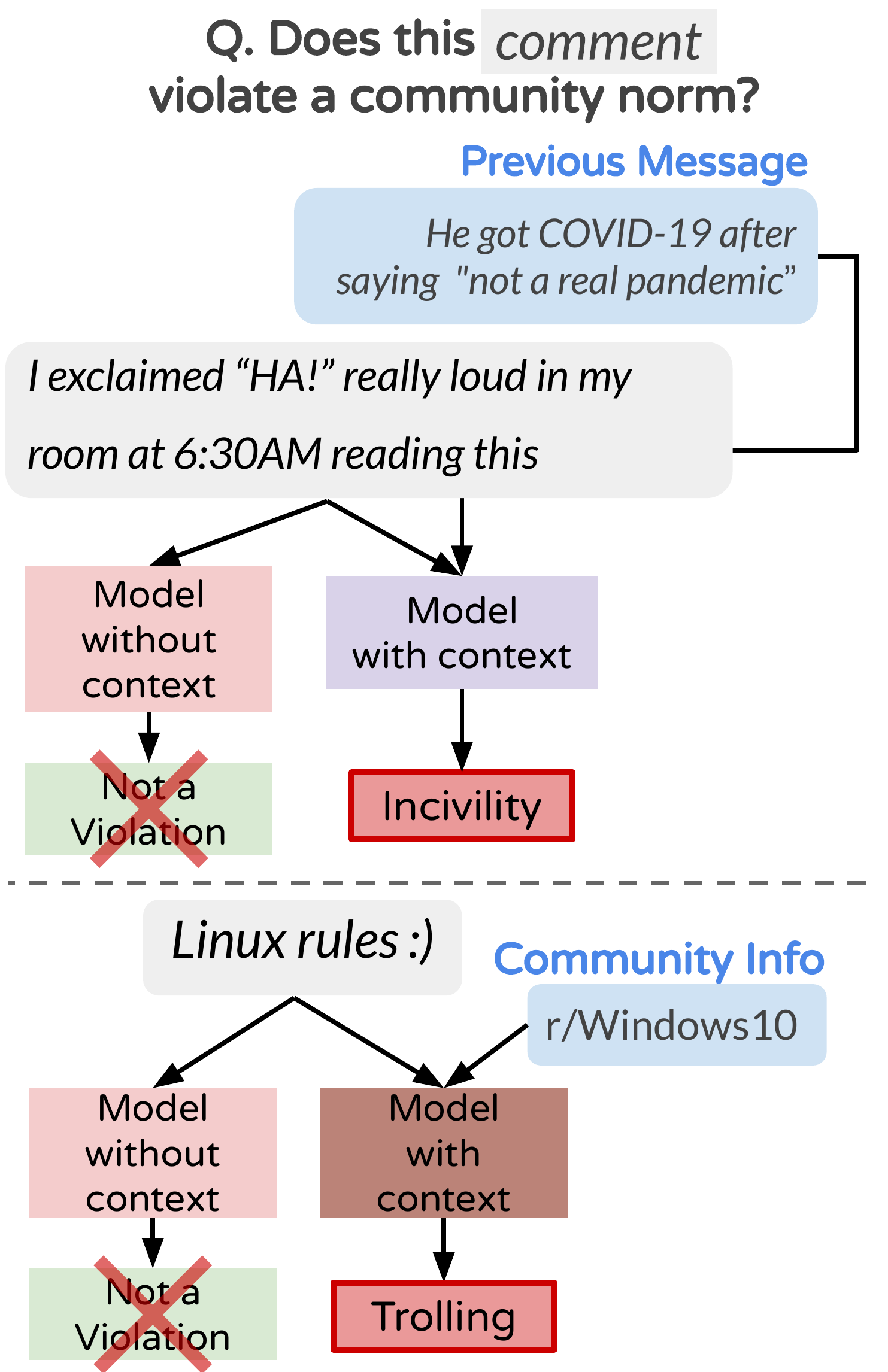}
  \caption{Two example comments\footnotemark{} that were moderated due to violating community norms. The examples highlight the importance of contexts (i.e. conversation history and community information) in detecting community norm violation.}
  \label{fig:motivation}
\end{figure}

\footnotetext{All example comments used in this paper are lightly paraphrased to preserve privacy.}

We introduce a new approach to context-sensitive automated content moderation that explicitly encodes community norms. Using a new dataset of 51K conversations across 3.2K communities, we show that the most commonly-studied norm violation behavior in NLP, hate speech, corresponds to a small minority of cases in which moderators intervene in practice. 
We then create multiple models to detect when moderators intervene and \textit{why} they intervene, adapting to the norms and rules of a community.

Our paper offers the following four contributions towards advancing the future of NLP in community and context-specific moderation.
First, in a large scale analysis of rule and moderation behavior, we show that subreddits vary considerably in their rules, with only some common themes. However, in practice, most rules are not enforced and, further, the enforcement of some types of rules, e.g., \textit{incivility}, is highly varied across communities. 
Second, we introduce a new dataset, \data{}, of 51K conversations across 3.2K subreddits and map the 25K rules from these communities into nine categories of context-specific unacceptable behavior, including five types of incivility. 
Third, we introduce a new series of models aimed at detecting and explaining rule-violating behavior based on norms and rules of the community. Our approach enables not only identifying that conversation in a particular community (with particular rules) is likely to violate a rule, but also \textit{which rule}. We demonstrate the effectiveness of these models, showing our best model attains an F1 of 78.64 across all rule types, a 50\% improvement over context-insensitive baselines. 
Finally, we perform an in-depth analysis of how much conversation context and community-sensitivity affects predictability. Our work points towards key challenges in detecting particular rule violations, while providing high accuracy in others, which can allow moderators to quickly intervene.
More generally, our work provides a clear next step for NLP to look beyond one-size-fits-all methods for detecting incivility to developing holistic, context-sensitive approaches that better suit the needs of moderators and their communities.

\section{\data Dataset}
\label{sec:data}
Prior work has created datasets  used to detect single types of norm violations in social media messages (e.g. incivility, hate speech or hostility) \citep{waseem-hovy-2016-hateful,founta2018large}. However, these datasets typically focus on isolated texts and do not provide prior conversational context or community-specific details.

In order to detect representative types of norm violations and account for context, we construct a new dataset---\data---a collection of 52K English conversation threads on Reddit.  \data includes comments removed for violating a variety of community norms beyond the traditional hate speech and incivility, such as spamming or violating community format/topics. Furthermore, \data provides additional context beyond the norm-violating comment itself with (a) the entire conversation thread (i.e., the original post and prior comments) and (b) the subreddit (i.e., community) in which the comment was posted.

\myparagraph{Data Collection}
We collected our initial data via the Reddit API, which provides list of moderators and their comments for each subreddit.
For each of the top 100K most popular subreddits,\footnote{Ranked by number of subscribers as of April 2021} we identified the most recent 500 comments from each moderator and retrieved comments that moderators posted in response to a removed comment (henceforth, \textit{moderation comments}). 

Moderation comments often provide useful signals for inferring which community norm was violated. From the full set of moderation comments, we selected those that contain a phrase explicitly stating the rule number 
(e.g.~``this comment violates Rule 2'') or the exact text of one of its subreddit's rules (e.g.~``don't be rude''). 

We then fetch the entire conversation thread for this set of moderation comments: the original post and all parent comments prior to the moderator's comment. We also fetched the norm-violating comment that was removed by moderators, by searching archived comments via the \href{https://pushshift.io}{Pushshift API} \citep{baumgartner2020pushshift}.\footnote{We were unable to retrieve an additional 21K removed norm-violating comments, which  were unavailable in the PushShift archive. We still include these corresponding conversations in our data release as they can be useful in the task of forecasting future norm violations.}

The final dataset is comprised of 20K conversations that have the last comment removed by one of the moderators of the community. 
Following the approach in \citet{chang-danescu-niculescu-mizil-2019-trouble}, we include 32K paired unmoderated conversations as a control set. 
Each moderated conversation is matched with up to two unmoderated conversations from the same post and with most similar conversation lengths as the target moderated conversation.

\myparagraph{Ethical Considerations for Protecting User Privacy}
Our dataset focuses, in part, on comments that moderators have viewed as objectionable and therefore removed. While these moderated comments are still publicly available, their use requires additional ethical reflection and precautions to preserve the dignity and privacy of users \citep{townsend2016social}. Moderated comments offer significant benefit to the study of supporting moderators and authorities in their goals of having supportive technologies that match their community's norms. At the same time, users who made those comments may object to having them included in a dataset \citep{fiesler2018participant}. Therefore, we  take additional measures to ensure that user privacy is protected, especially for the deleted comments. 

We use Reddit data through Pushshift \cite{baumgartner2020pushshift}, an archive that has been widely used in NLP and related fields since its first release in 2015 \cite[\textit{among many others}]{hessel-lee-2019-somethings,kennedy-etal-2020-contextualizing, sap-etal-2020-social,dinan-etal-2020-queens}. 
Pushshift's collection policy explicitly states that it conforms to Reddit's rules and user agreement with regards to data collection. 
In releasing our dataset, we provide only the associated identifiers of comments but \textit{not} their textual content. Practitioners will need to independently fetch the texts from Pushshift by using the provided comment IDs. Releasing only IDs ensures that any users who request their data to be removed in Pushshift will also have it removed in our dataset.
Additionally, in our dataset we anonymize individual usernames and personal identifiers of posters and moderators.
Finally, along with our data release, we provide  guidelines to the users who wish to delete their comments from the Pushshift dump.

\myparagraph{Classification of Community Norms}
Moderator comments as well as rules defined in each subreddit are free-form and diverse, and it is not trivial to map the rule/comment to a specific community norm it refers to. In order to study norm violations, we thus first train classifiers that given a rule description label it with a type of norm it violates. 

We follow \citet{Fiesler_Jiang_McCann_Frye_Brubaker_2018}'s qualitative analysis of 1K subreddits, that identified main categories of rules through annotating 3,789 rules from the subreddits.\footnote{Out of 24 categories, we exclude the ones describing the tone of rules (whether a rule is ``Prescriptive'' or ``Restrictive'') and one (Behavior/Content) that is extremely broad, covering over 90\% of coded rules.} We then use the annotations from \citep{Fiesler_Jiang_McCann_Frye_Brubaker_2018} to fine-tune a BERT-based binary classifier for each rule type.\footnote{Binary classifiers were used since each community rule can be categorized with multiple types. We used the default hyperparameters suggested in the \href{https://github.com/huggingface/transformers}{Transformers} library and trained each model for 20 epochs.} 
\Tref{tab:rule-classifier} shows the list of the resulting 21 categories of community norms and the performance of our classifiers evaluated using macro F1 scores with stratified 10-fold cross validation.

\begin{table}[t]
\centering
\resizebox{0.85\columnwidth}{!}{%
\begin{tabular}{llll}

\textbf{Rule Types}                           & \textbf{F1}   & \textbf{Rule Types}     & \textbf{F1}   \\
\hline
Advertising       & 71.0 & NSFW          & 88.2 \\
\begin{tabular}[c]{@{}l@{}}Moderation\\ Enforcement\end{tabular}& 87.0  & Off-topic     & 63.5 \\
Copyright/Piracy                    & 70.6 & \begin{tabular}[c]{@{}l@{}}Personal\\ Army\end{tabular} & 43.2 \\
Doxxing               & 75.4 & Personality   & 81.9 \\
Format                              & 73.5 & Politics      & 85.7 \\
Harassment                          &  67.9     & Reddiquette   & 83.2 \\
Hate Speech                         & 84.2 & Reposting     & 81.4 \\
Images                              & 65.1  & Spam          & 86.9 \\
Outside Content              & 68.0 & Spoilers      & 76.7 \\
\begin{tabular}[c]{@{}l@{}}Low-Quality\\ Content\end{tabular} & 45.6 & Trolling      & 96.0 \\
&& Voting &85.6\\
\midrule
AVERAGE & 75.3 & &\\
\end{tabular}
}
\caption{Macro F1 of classifying the diverse sets of rules across subreddits to  rule violation types.}
\label{tab:rule-classifier}
\end{table}

We use the final models to map 183K rules from the top 100K subreddits to their corresponding rule types. 
\Tref{tab:rule_types} shows the examples of labeled community rules randomly sampled from our data.
Finally, we classify moderators' explanations of the rule-violating comments in \data. Because we only kept moderators' comments that mention a rule number or a rule's exact text, we can determine which rule was violated by the conversation.\footnote{Any data collection procedure that relies on user-generated labels has the risk to absorb human biases. In our setting too, there is a risk of moderator biases to be incorporated when we match moderation comments to rules and violation types. However, in pilot work examining moderator comments with explicit rule violations and those where we had to infer the rule(s), we found a near-identical distribution of violation types.} Using our binary classifiers on rule text, we can now infer the type of norm that was violated by the moderated (removed) comment. 

Although the 21 types are well suited for fine-grained analysis of rules on Reddit, they might leave insufficient number of examples per type which can make it more challenging to computationally model them.
We define relatively more coarse-grained nine types and map the 21 types with the nine types as shown in  \Tref{tab:rule_types}.
We designed these types to reflect our interest in text-based analysis of abusive language.
We kept five different subcategories of uncivil comments (general incivility, trolling, harassment, hate speech, spam) while aggregating Voting, Reddiquette, and Moderation Enforcement into a broad "Meta-rules"  category. 
In the remainder of this paper, we only use the coarse-level norm violation types.

Ultimately, each moderated comment in \data has the following information: (1) its subreddit, (2) its conversation thread, (3) the community-specific rule violated, and (4) the coarse- and fine-grained rule types that were violated. To maximize user privacy, all comments are provided as IDs, the content for which can be retrieved through the Reddit and PushShift APIs.

\begin{table}[]
    \centering
    \resizebox{\columnwidth}{!}{%
    \begin{tabular}{lr}
    \toprule
    \textbf{Incivility}: \{Personality\}                       & \textit{``Be civil"}                    \\
    \midrule
    \begin{tabular}[c]{@{}l@{}}\textbf{Harassment}: \{Harassment, \\Doxxing\}\end{tabular}                        & \textit{``Don't harass others"}                             \\
    \midrule
    \begin{tabular}[c]{@{}l@{}}\textbf{Spam}:\{Spam, Reposting, \\Copyright\}\end{tabular}                         &      \textit{``No excessive posting"}                       \\
    \midrule
    \begin{tabular}[c]{@{}l@{}}\textbf{Format}: \{Format, Images,\\ Links\}\end{tabular}                            &           \textit{``Use the correct tags''}                  \\
    \midrule
    \begin{tabular}[c]{@{}l@{}}\textbf{Content}:\{Low-quality Content,\\ NSFW, Spoilers\}\end{tabular}                           & \textit{``No low-quality posts''}                             \\
    \midrule
    \textbf{Off-topic }:\{Off-topic, Politics\}                        & \textit{``Only relevant posts''}                             \\
    \midrule
    \textbf{Hate speech}:\{Hatespeech\}                       & \textit{``No racism, sexism''}                             \\
    \midrule
    \begin{tabular}[c]{@{}l@{}}\textbf{Trolling}:\{Trolling, \\Personal Army\}\end{tabular}                       & \textit{``No trolls or bots''}                             \\
    \midrule
    \begin{tabular}[c]{@{}l@{}}\textbf{Meta-rules}:\{Voting, Moderation \\ Enforcement, Reddiquette\}\end{tabular}                     &     \textit{``No Downvoting''}                       \\
    \bottomrule
    \end{tabular}
    }
    \caption{The mapping between coarse- and fine-grained rule types and examples.}
    \label{tab:rule_types}
\end{table}

\begin{figure}[t]%
    \centering
    \includegraphics[width=\linewidth]{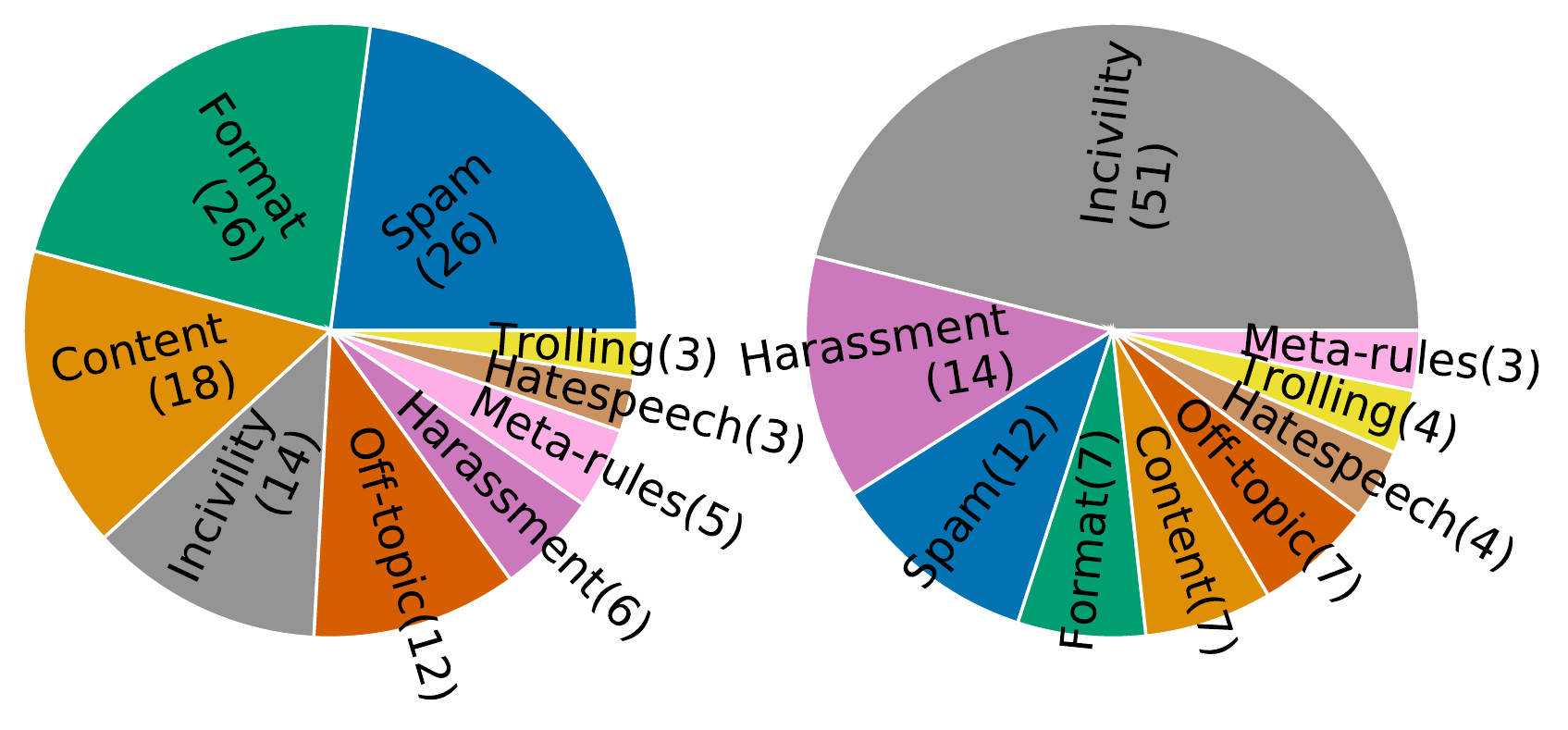}
    \caption{\% of rule types of rules (left) and comments violating those rules (right) in \data.}%
    \label{fig:proportion}%
\end{figure}

\begin{figure}[t]
    \centering
    \includegraphics[width=\linewidth]{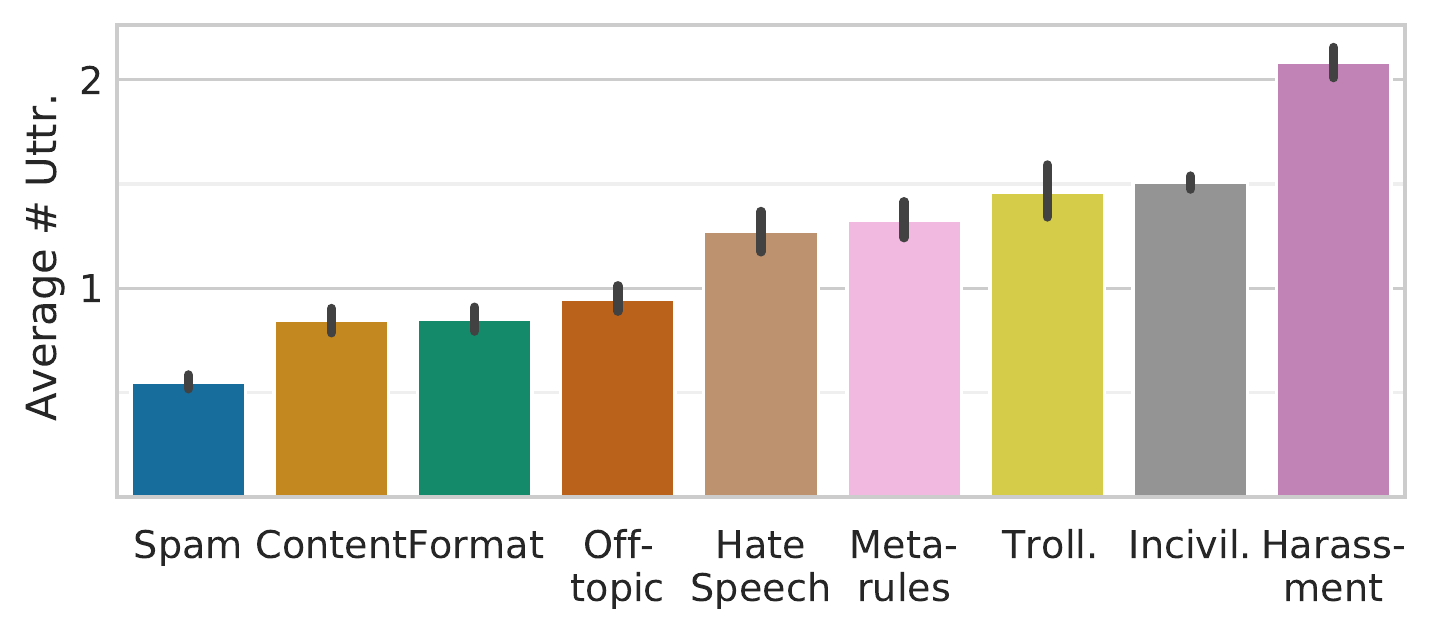}
    \caption{Average number of utterances between the original post and the moderated comments.}
    \label{fig:num_uttr}
\end{figure}

\myparagraph{Analysis of Community Norm Violations}
We analyze the types of rules and comments comprising \data with a focus on what kinds of rules are established by communities, what kinds of rules are violated in practice, and when in conversations these rules are violated.

The results in \Fref{fig:proportion} show that the rule types are evenly distributed over rules (left) while the actual violations (right) are relatively more focused on abusive language rule types such as Incivility and Harassment.
A large proportion of all rules in our dataset fall under the Format and Content categories, suggesting that there is a diverse set of community norms, beyond regulating incivility, needed to operate healthy online communities. 
Critically, while the majority of efforts on identifying abusive language in the NLP community have been focused on hate speech, more subtle types of incivility are significantly more prevalent in removed comments, which are also harder to detect \cite{jurgens-etal-2019-just,breitfeller2019finding,field-tsvetkov-2020-unsupervised}. Moreover, only 55\% of removed comments are violations of Incivility and Hate Speech rules, again highlighting the importance of understanding the spectrum of community norms in designing automated moderation assistance systems.

\Fref{fig:num_uttr} shows the average number of utterances from the original post to the norm-violating removed comment. Overall, violations related to abusive language such as Harassment, Incivility, and Trolling occur \textit{later} in conversations than comments removed for other reasons (e.g. Spam and Format).
This timing has implications for the ``forecastability'' of violation types. For example, the average conversation length within the Spam category is about 0.5 which indicates that half of the violations happen in the original post or a reply to it, making it impractical trying to forecast such violations.

Even though Hate Speech and Harassment are both related to abusive language, comments removed due to Harassment occur after more interactions. We hypothesize this is because harassment and trolling are intentionally expressed in less overt forms to delay the moderators' intervention.
These findings illustrate that with a more representative set of community rules and a larger-scale dataset, \data facilitates deeper understanding of community norm violation behaviors and provides guidance on more urgent tasks our field should be focusing on for a practical impact.

\begin{figure*}[t!]%
    \centering
    \subfloat
    {{\includegraphics[width=0.45\linewidth]{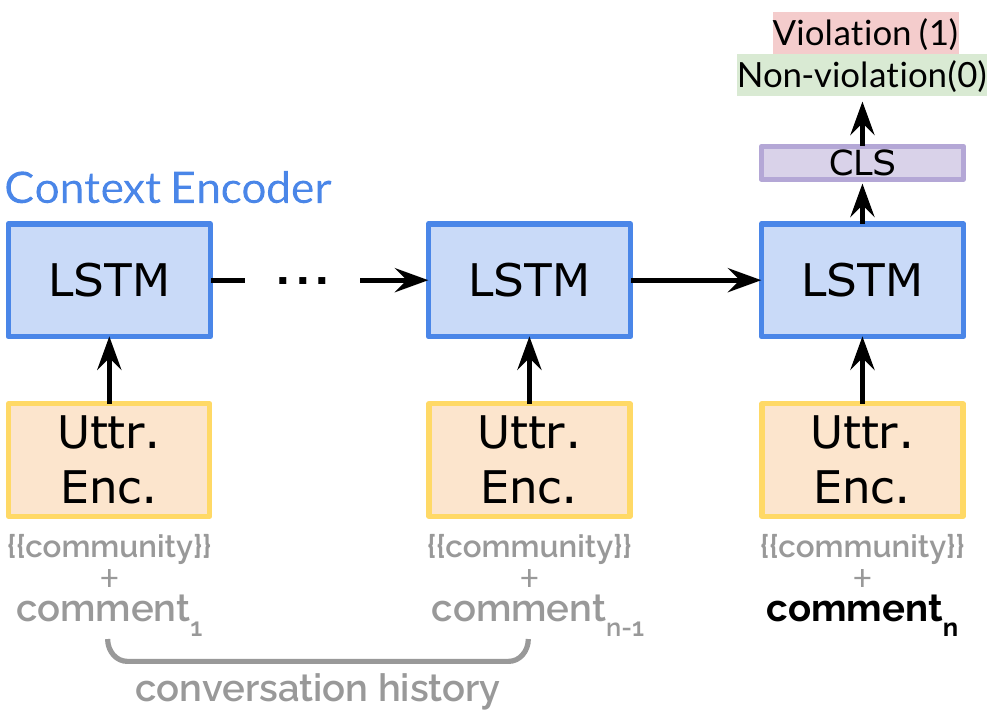} }}%
    \hfill
    \subfloat
    {{\includegraphics[width=0.45\linewidth]{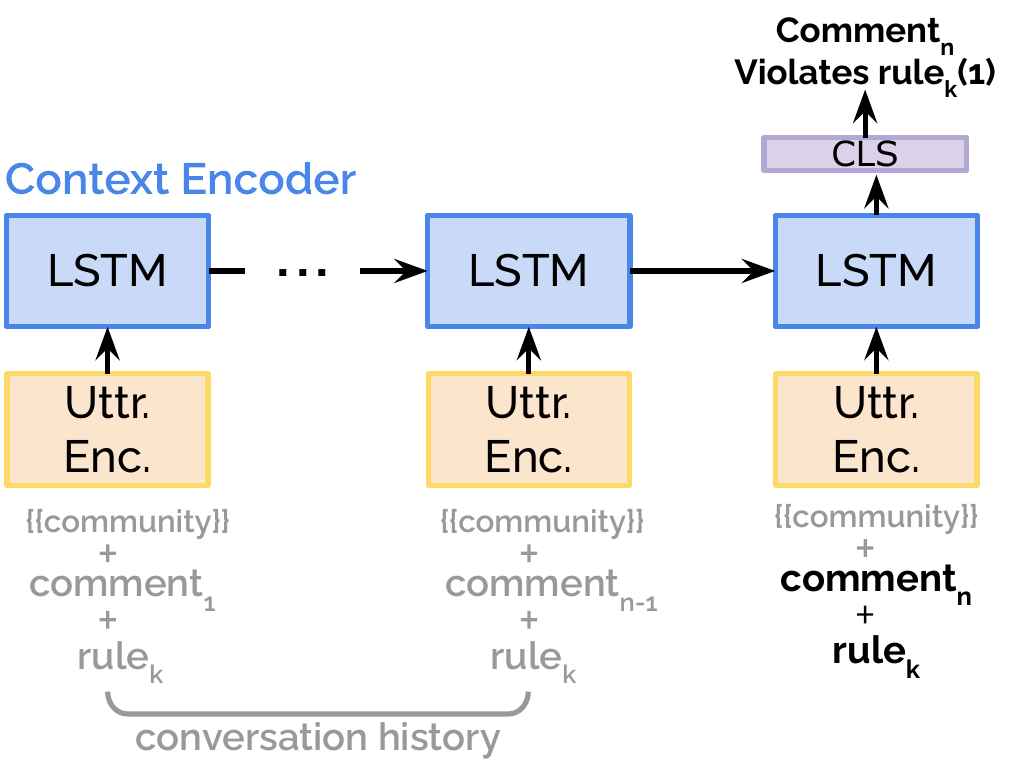} }}%
    \caption{Structure of the baselines of the two proposed tasks: detecting norm violation (left) and explaining norm violation (right).
    Inputs  in gray (conversation history and community information) are optional context.}%
    \label{fig:model}%
\end{figure*}

\section{Detecting and Explaining Community Norm Violations}
\label{sec:models}
With \data, we can now train models for detecting contextualized, fine-grained community norm violations. 
We present two tasks: (1) Detecting community norm violations, and (2) Explaining community norm violations. The former identifies coarse categories of norm violations detailed in \Sref{sec:data}, and the latter is aimed at identifying specific local community rules being violated, to facilitate moderation transparency.  
For each task, we compare model variants without or with varying types of incorporated context, including conversation history and community information (e.g.~subreddit name).

\subsection{Detecting Community Norm Violations}
\label{sec:norm-violation}
In this task we assume a set of pre-defined categories of norm violations. 
For each category, we train a binary classifier to detect violations, since the categories are not mutually exclusive.

As shown in \Fref{fig:model}, we encode a conversational context of arbitrary length along with community rules. Following \citet{chang-danescu-niculescu-mizil-2019-trouble}, we use a uni-directional LSTM context encoder. The utterance encoder is initialized with a pretrained BERT model, with each classifier is then fine-tuned using training data specific to each rule type (data statistics are detailed in \Aref{sec:data-description}). The last hidden state from the last comment is fed into the classifier. The flexibility of this design allows for both retroactive detection after violations occur (the focus of this work) as well as proactive prediction of future rule violations.

We experiment with four model variants with different input contexts: 
\begin{itemize}[leftmargin=*,topsep=0pt]
    \setlength{\itemsep}{-4pt}
    \item {\finalcomment{}}: Only the final comment.
    \item {\history{}}: Past conversation history and the final comment.
    \item {\community{}}: Community information and the final comment. We concatenated the subreddit name in front of the comment (e.g. ``r/AskReddit ask anything!'').\footnote{Note that the model variants without conversation history do not use a context encoder at all and thus have a smaller number of trainable parameters.}
    \item {\histcommunity{}}: Conversation history and community information. 
\end{itemize}

\subsection{Explaining Community Rule Violations}
\label{sec:rule-violation} 
In addition to categorizing rule violations by type (type-based), we develop a model that leverages the specific community rule text to identify violations in context. This text-based model facilitates explanations of rule violations, and improves transparency \citep{juneja2020through}. Such a system could lighten moderators' workload through highlighting why they might moderate a comment, enable more productive interventions, and improve the relationship between community members and moderators. 

Similar to the violation category detection task, we construct binary classifiers that detect violations given conversational and community context. However, as shown in \Fref{fig:model}, the full input and training procedure
are different; we include the community's verbatim rule description as a model input. 
The rule text is appended to the input comment with
a special token ([SEP]) added between the comment and the rule to leverage pretrained language models' ability to infer relationships between two sentences.
Since the precise formulation of the target rule is given as an input, we no longer need to train one model per rule type; we train one universal model with all available training data.

\data contains information about which rules are violated in each removed comment, and we use these rule-comment pairs as positive examples. If a comment is tagged for violating more than one rule, we include all comment-rule pairs as positive examples. We construct negative training examples using matched unmoderated conversations from \data (described in \Sref{sec:data}) by adding the text of the violated rule to the corresponding unmoderated conversation. 

To guide the model in better discriminating rules, 
we construct additional negative examples by mapping each removed comment with an randomly chosen incorrect rule from the same subreddit (e.g. ``Here's my referral code! [SEP] No Politics'').

Similarly, we experiment with three model variants with different input contexts:
\begin{itemize}[leftmargin=*,topsep=0pt]
    \setlength{\itemsep}{0em}
    \item {\ruletext}: Only the final comment and a rule text.
    \item {\rulehist}: Past conversation history, the final comment, and a rule text. 
    \item {\rulecommunity}: Both conversation and community history, the final comment, and a rule text.
\end{itemize}

The main advantage of the text-based model is in its interpretability and generalizability. 
Since the model now looks at the community-specific rule texts, the system can provide more meaningful feedback to moderators and users. 
For example, instead of saying ``potential hate speech detected'', now the model can be more informative in notifying users that ``the comment has breached our community's Rule 2: No Racial Slurs''. 
Moreover, since the model takes free-form rules as input, it can generalize to unseen rules and novel rule types.

\section{Experiments}
\label{sec:experiments}
\begin{figure*}[t]
    \centering
    \includegraphics[width=\linewidth]{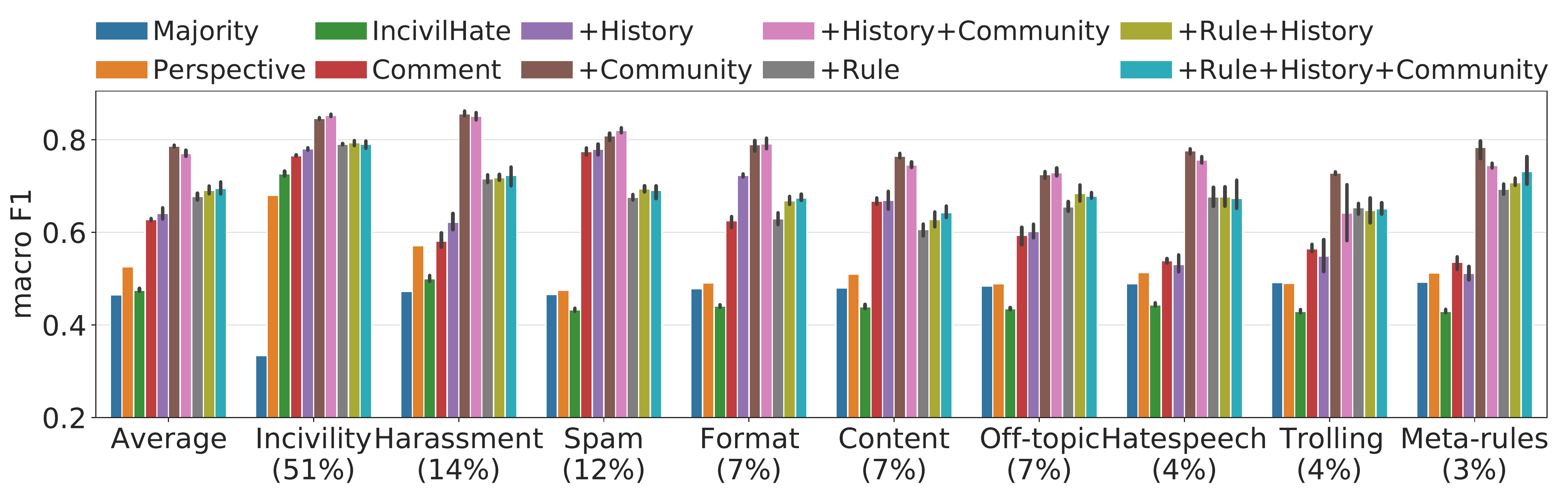}
    \caption{Average and breakdown of Macro F1 scores of the baselines and the model variants. 
    Error bars indicate 95\% confidence interval and the types are sorted by their violation frequency (percentage below x-axis labels). 
    }
    \label{fig:model_res}
\end{figure*}

\myparagraph{Baselines}
In addition to the seven model variants in \Sref{sec:models}, we consider three baselines that represent current common approaches:
\begin{itemize}[leftmargin=*,topsep=0pt]
    \setlength{\itemsep}{0em}
    \item \majority: Majority class baseline.
    \item \perspective: \href{https://www.perspectiveapi.com}{Perspective API}'s toxicity score of the final comment to make a binary decision. For each rule type, a threshold value was tuned to maximize development set F1 score.
    \item \incivil: We train a model using just the incivility and hate speech violations from \data. The test set predictions from the trained model was evaluated over different rule types.
\end{itemize}

\myparagraph{Training Details}
We perform an 80-10-10 train/dev/test random split of moderated comments in \data{} and then 
appended paired unmoderated comments into the same split.
The resulting number of examples of train/dev/test split was 41667, 5214, and 5131, respectively. We ran training for five different random seeds and report the average scores of multiple runs except for \majority{} and \perspective{} baselines. 

The base utterance encoder is a pretrained   \href{https://huggingface.cfo/DeepPavlov/bert-base-cased-conversational}{Conversational BERT} model. 
Each model was trained for 10 epochs with an early stopping patience of 5, and with Adam optimizer with a learning rate of 1e-5.
We used a batch size of 32 for models that do not leverage past conversations and 8 for the ones that use comment history. 
We used 2 layers of GRUs with a hidden size of 768 for the context encoder and 2 linear layers for the final classifier.

\myparagraph{Evaluation}
We used macro F1 to evaluate all models.   
For models in \Sref{sec:rule-violation}, at test time we cannot assume that we know which rule will be violated in a given conversation.
We thus create multiple comment-rule pairs for each comment in the test set by matching it with each community rule. 
Out of the resulting pairs, we mark the pairs that were observed in the original test set as positive, and the remaining pairs are marked as negative. We refer to these negative pairs added to the test set of models explaining rule violations as \textit{augmented pairs}.
Note that the test sets of models in \Sref{sec:rule-violation} are now different from the text sets in \Sref{sec:norm-violation} and the F1 scores of two tasks are not directly comparable.

\myparagraph{Experiment Results}
Information from the social context of a comment substantially improves performance (\Fref{fig:model_res}). Compared to current approaches for inferring toxicity, all type-based violation detection model performed significantly better---even for rule violation categories those approaches are tailored for. While \perspective{} and \incivil{} performed better in Incivility and almost comparable with \finalcomment{} and \history{}, adding community information still resulted in a significant improvement of +8.0 absolute increase in F1.

Across all rule violation types, adding the context about community  significantly improved the performance, often resulting in the highest performing models when added.
Adding conversation history showed mixed results. 
\history{} showed improvements over \finalcomment{} whereas \histcommunity{} was not necessarily better than \community.
Models with conversation history tend to perform worse on scarce violation types such as Meta-rules and Trolling; we speculate that this decreased performance is due to the increased number of parameters from adding context encoder layer to process conversation history and future work with more examples of these violations may substantially improve performance.
This result for history greatly expands an analysis by \citet{pavlopoulos-etal-2020-toxicity} that found minimal performance gain when adding a single prior comment to identify toxicity; while we too find minimal improvement for Incivility and Harassment norms, adding history \textit{does} improve the recognition for other norm violations (e.g., Format and Content) indicating that prior context can be useful. 

While the results of text-based violation detection models (\ruletext{}, \rulehist{}, \rulecommunity{}) and type-based models are not directly comparable due to the augmented pairs, they were evaluated over the same set of comments so the numbers can provide a general sense of text-based model performance. 
An interesting distinction between the two detection tasks is in how much additional context helps.
In type-based models, adding context made significant improvements in all or in some cases. However, with text-based models, the performance was relatively more uniform and additional context did not contribute as much. This result suggests that providing full text of rules may help resolve certain ambiguous comments and thus the model rely less on the additional context.

\section{Analysis}
\label{sec:analysis}
\begin{figure}[t]
    \centering
    \includegraphics[width=0.78\columnwidth]{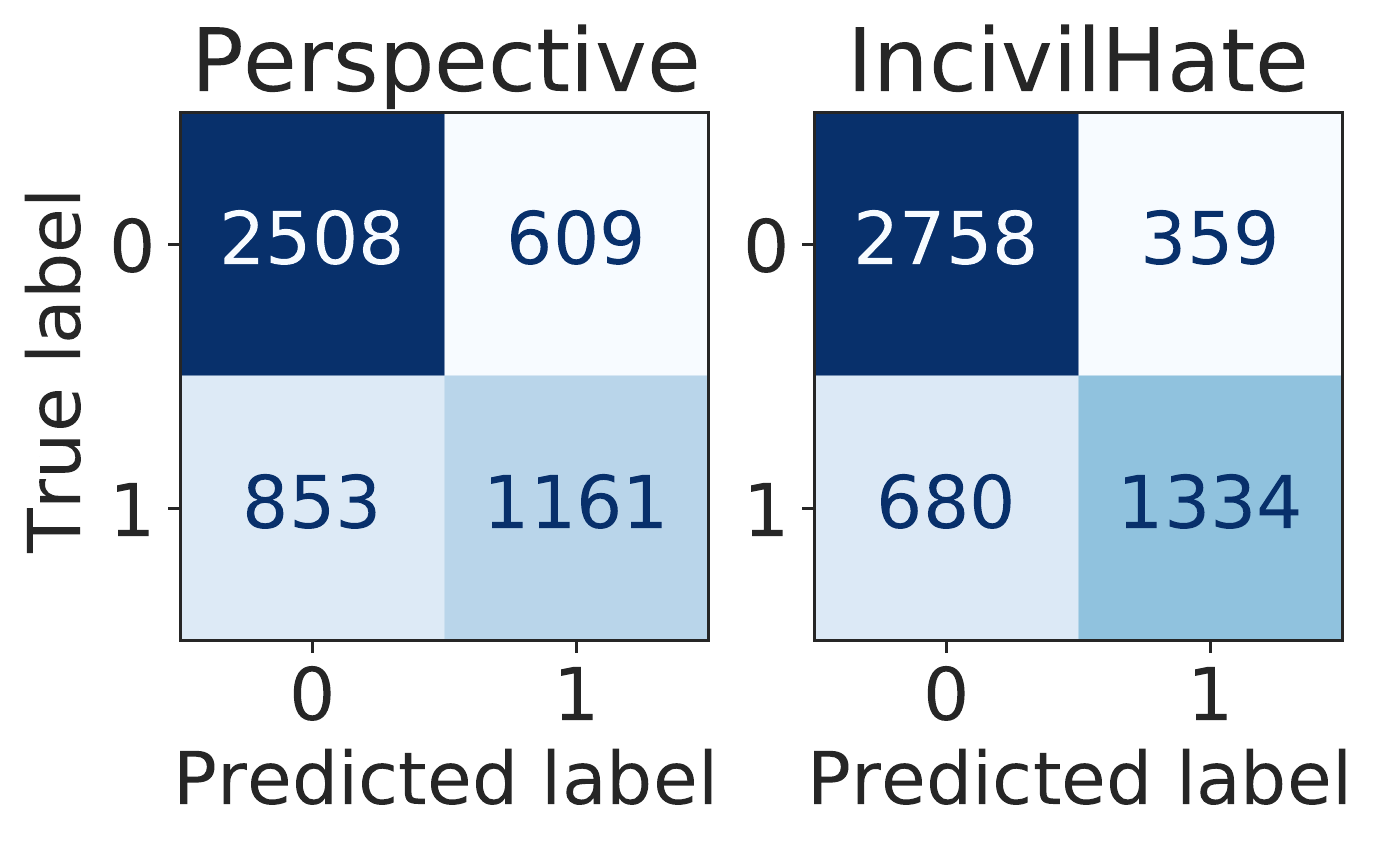}
    \caption{Confusion matrices of two baselines over the norm violation detection task.}
    \label{fig:cm_baseline}
\end{figure}

\begin{figure}[t]
    \centering
    \includegraphics[width=0.8\columnwidth]{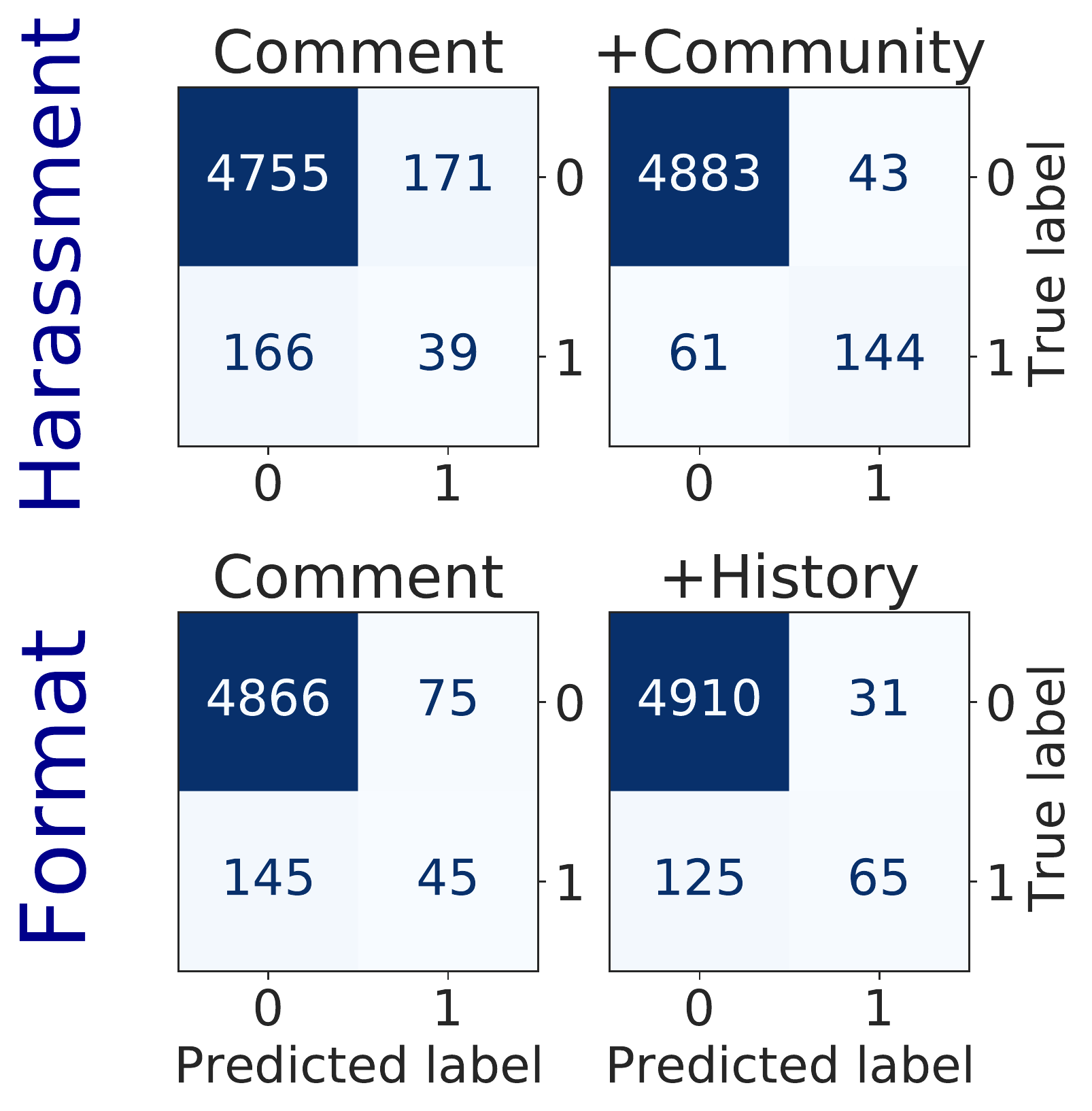}
    \caption{Confusion matrices of \finalcomment{} and \community{} for the Harassment detection task (top), and of \finalcomment{} and \history{} for the Format violation detection task (bottom).
    }
    \label{fig:cm_combined}
\end{figure}
\myparagraph{How many violations do current systems miss?}
In part due to their targeted focus, the \perspective{} and \incivil{} baseline models miss a substantial proportion of the total norm violations. \Fref{fig:cm_baseline} shows the confusion matrices of the violation detection task, where labels are aggregated over all violation types to test how many violations overall are not captured by these systems. 
The results show that \perspective{} and \incivil{} fail to recognize 42\% and 34\% of all violations, respectively. 
Moderators on platforms like Reddit must triage huge numbers of comments daily and this points to a clear gap between current practice (represented by the baselines) and indicates what moderators act on in practice.

\myparagraph{How does community information help?}
We observed that adding community information provides the most significant improvements in Harassment in \Fref{fig:model_res}.
We now look into the Harassment type to understand more about how did the additional community information actually help to improve the performance.

What kinds of errors are corrected by adding community context? By comparing confusion matrices of \finalcomment{} and \community{} (\Fref{fig:cm_combined}), we find that \community{} has fewer false positives.
Out of 154 false positives from the \finalcomment{} model that were corrected in the \community{} model, 106 (69\%) were Incivility violations. 
Consider the following example:

{
\centering
\begin{tcolorbox}[width=0.95\columnwidth, colframe=white!68!black, boxsep=1pt,left=0.5pt,right=0.5pt,top=0.5pt,bottom=0.5pt]

\textbf{Comment}: \\
\textit{That game's already dead to 99\% of the world a few weeks later, get over it you stupid idiot.}\\
\textbf{Moderator Comment}: \\
\textit{Your comment has been removed for Rule 2. Be civil and respectful. Do not attack or harass other users or engage in hate-speech.}\\
\textbf{Paired Rule}: Rule 2: Be civil and respectful.\\
\textbf{Violation}: Incivility\\
\textbf{Community}: r/classicwow
\end{tcolorbox}
}

The final comment in this example could be considered as both a Incivility and Harassment violation and \finalcomment{} model labels it as Harassment. Although the moderator refers to the community's Incivility rule, the rule mentions "do not attack or \textit{harass} other users", which makes it clear that this example falls into both categories. However, the \community{} model labels this comment as Incivility and not Harassment. We speculate that the \community{} model learns about what rules exist in each community; \textit{r/classicwow} has 8 rules and none of them are about Harassment, so moderators refer to the Incivility rule when moderating Harassment violations. 
In other words, depending on the community and their available community rules, the same comments can be moderated as either incivility or harassment violation. 
Therefore, providing the community information can help the model disambiguate this decision and ground its moderator support in the norms of the community.

\myparagraph{How does conversation history help?}
Likewise, for the conversation history context, the largest gain was achieved in the Format type.
In \Fref{fig:cm_combined}, we compare confusion matrices of \finalcomment{} and \history{}. The result again shows that additional context can help the model in reducing the false positive rate.

Among the corrected false positives, the most prevalent type mistaken for Format was Spam.
One example of such case is given below: 

{
\centering
\begin{tcolorbox}[width=0.95\columnwidth, colframe=white!68!black, boxsep=1pt,left=0.5pt,right=0.5pt,top=0.5pt,bottom=0.5pt]
\textbf{Comment}: \textit{UPDATE: I found it! here you go if you need it\_LINK\_\\}
\textbf{Violation}: Spam (Piracy)\\
\textbf{Moderator Comment}: \\
\textit{See Rule 1: No Merchandise / Spam}\\
\textbf{Previous Message}: \\
Does anyone know where to buy this?
\end{tcolorbox}
}

If we only consider the final comment, there are two possible explanations for which rule was violated: 1) Format: the outside link does not follow the community guideline 2) Spam: self-promotion / promoting specific merchandise is banned. 
However, the previous message makes it clear that the author had just posted about a product and then made a self-reply with a link to buy the product.
With this information, model can disambiguate this situation  and choose the right violation type.

\section{Related Work}
\paragraph{Community Norms and Rules}

Many studies have investigated how online conversations are moderated and how each community has different norms to ensure a safe environment for discussions \cite{chandrasekharan2018internet,jhaver2018online,jhaver2019human,juneja2020through, almerekhi2020investigating,rajadesingan2020quick}. 
\citet{Fiesler_Jiang_McCann_Frye_Brubaker_2018} conduct an analysis over the rules of Reddit communities and define 24 types of the rules. 
They provide a thorough and large-scale analysis over how the rules are phrased and how rules are different across subreddits. 
We adopt their rule categorization and extend it to code actual rule violations. 

\citet{chandrasekharan2018internet} also studied removed comments on Reddit to understand what types of rules exist on Reddit by clustering the moderator comments and investigated how they are governed. 
However, their dataset provides limited context of moderated comments, whereas we focus on providing a dataset that has enough context and also explicit violation type that can be leveraged in modeling rule violation.

\myparagraph{Context in Detecting Online Abuse}
Most of the existing datasets for abusive language detection implicitly assumes that comments may be judged independently taken out of context. 
\citet{pavlopoulos-etal-2020-toxicity} challenged this assumption and examined if context matters in toxic language detection. 
While they found a significant number of human annotation labels were changed when context is additionally given, they could not find evidence that context actually improves the performance of classifiers.
Our work also examines the importance of context, but we do not limit our scope to toxic language detection and investigate a broader set of community norm violation ranging from formatting issues to trolling.

\myparagraph{Beyond Incivility and Hate Speech}
\citet{jurgens-etal-2019-just} claims ``abusive behavior online falls along a spectrum, and current approaches focus only on a narrow range'' and urges to expand the scope of problems in online abuse.
Most work on online conversation has been focused on certain types of rule violation such as incivility and toxic language \citep[e.g.,][]{zhang-etal-2018-conversations, chang-danescu-niculescu-mizil-2019-trouble, almerekhi2020investigating}.
In this work, we focus on a broader concept of \textit{community norm violation} and provide a new dataset and tasks to facilitate future research in this direction.

\section{Conclusion}
\label{sec:conclusion}

Online communities establish their own norms for what is acceptable behavior. However, current NLP methods for identifying unacceptable behavior have largely overlooked the context in which comments are made, and, moreover, have focused on a relatively small set of unacceptable behaviors such as incivility. In this work, we introduce a new dataset, \data{}, of 51K conversations grounded with community-specific judgements of which rule is violated. Using this data, we develop new models for detecting context-sensitive rule violations, demonstrating that across nine categories of rules, by incorporating community and conversation history as context, our best model provides a nearly 50\% improvement over context-insensitive baselines; further, we show that using our models, we can \textit{explain} which rule is violated, providing a key assistive technology for helping moderators identify content not appropriate to their specific community and better communicate to users why.
Our work enables a critical new direction for NLP to develop holistic, context-sensitive approaches that support the needs of moderators and communities.

\section{Ethical Considerations}
We hope to draw attention to the mismatch between the standard tasks of  harmful content detection that NLP researchers are typically focusing on (e.g.~sentence-level toxicity detection) and the broad spectrum of context-sensitive content violation types that actually occur in the wild. 
To enable future research on detecting community-specific norm violations, we constructed a dataset that retrieves online conversation threads and comments deleted by moderators, categorized by community norm violations. 
We discuss ethical considerations related to protecting user privacy in \Sref{sec:data}.

Additionally, we acknowledge that the dataset itself can incorporate unintentional biases. For example, it can incorporate moderators' biases in deciding which comments are selected to be removed \cite{binns2017like,myers2018censored,shen2019discourse}. The unmoderated comments can include norm-violating comments that were missed by the moderators \cite{chandrasekharan2018internet}. By constructing a large scale dataset that spans multiple subreddits and moderators' teams we partially mitigate these concerns. To investigate this further, future work could incorporate an additional evaluation procedure with test sets containing held-out moderators  \cite[cf.][]{geva2019we}.

\newpage
\section*{Acknowledgments}
This material is based upon work funded by the DARPA CMO under Contract No.~HR001120C0124, and by the National Science Foundation under Grants No.~IIS2040926 and 1850221. The views and opinions of authors expressed herein do not necessarily state or reflect those of the United States Government or any agency thereof.

\bibliographystyle{acl_natbib}
\bibliography{anthology,derailment_ribiber}

\clearpage

\appendix
\section{Dataset Description}
\label{sec:data-description}

\begin{table}[h!]
    \centering
    \resizebox{\columnwidth}{!}{
    \begin{tabular}{lr}
    \toprule
\textbf{\data{}}&\\
\# of total comments                       &  52012\\
\begin{tabular}[c]{@{}l@{}}\quad\# of moderated comments with\\\quad original final comment restored\end{tabular}  & 20137 \\
\quad\# of unmoderated comments                                                                                  &  31875\\
\midrule
\textbf{Additional dataset for forecasting}&\\
(without original violation comments)&\\
\# of total comments                       &  53829\\
\# of moderated comments         &  20727 \\
\# of unmoderated comments &  33102 \\
\midrule
    \# of subreddits                                                                                              & 3234 \\
    \# of rules                                                                                                   & 24916 \\
    \# of moderators                                                                                              & 29841 \\
    \# of moderators per subreddit                                                                                & 9.2 \\
\midrule
    Avg. comment length (\# of words)                                                                          & 34.4 \\
    \begin{tabular}[c]{@{}l@{}}Avg. number of context per comment\\ (including the original post)\end{tabular} &  2.8\\
    Avg. \# of rules per community                                                                             & 7.7\\
    \bottomrule
    \end{tabular}
    }
    \caption{Summary statistics of \data}
    \label{tab:data-description}
\end{table}

\Tref{tab:data-description} presents the basic summary statistics of \data.
Our main dataset used in the analysis consist of 52K comments in total, and each comment is accompanied with its conversation history, subreddit information, tagged rule, and its violation type.

The dataset also provides additional 54K comments that contains 21K violation comments and its paired 33K unmoderated comments. For these moderated comments, we could not fetch its original comment before getting moderated, so these could not be used for detection task.
However, these comments could still be used in training norm violation \textit{forecasting} models.

\section{Additional Details for Reproducibility}
Our work includes two series of model training: rule classifier training and violation detection model training. 
For all training runs we trained with one GPU with 11GB of memory.

For rule classifiers, we had to train one binary model for each violation type, so we had to run 21 final training using 3.7K annotated rules. Each run took about 5-6 minutes which results in about 2 hours of training. 

Violation detection models are trained with 52K examples thus took significantly longer than training rule classifiers.
Again, for type-based detection models, we needed to train one model per coarse-grained violation types. Each run took about 40 minutes for models without conversation history and took about 2 hours for models with history. 
In summary, to run one set of training for one model, we needed to train for 6 hours for models without history and 18 hours for models with history.

For text-based detection models, we did not need to train a model per type which significantly reduces the total training amount.
Models without conversation history took about an hour to train and models with history took about 7 hours to train one model. 

The number of trainable parameters was 109 million for models without conversation history (i.e., those without a context encoder) and 116 million for models with a context encoder.

\end{document}